\documentclass[conference]{IEEEtran}
\IEEEoverridecommandlockouts
\newcommand{\paperversion}{3}
\ifcase\paperversion
  \newcommand{\versionnotice}{}
\or
  \newcommand{\versionnotice}{\thanks{This work has been submitted to the IEEE for possible publication. Copyright
  may be transferred without notice, after which this version may no longer be accessible.}}
\or
  \newcommand{\versionnotice}{}
\or
  \newcommand{\versionnotice}{\thanks{\copyright~2026 IEEE. Personal use of this material is permitted. Permission
  from IEEE must be obtained for all other uses, in any current or future media, including reprinting/republishing
  this material for advertising or promotional purposes, creating new collective works, for resale or
  redistribution to servers or lists, or reuse of any copyrighted component of this work in other works.}}
\fi
\newif\ifwithappendix
\ifnum\paperversion=1 \withappendixtrue\fi
\ifnum\paperversion=3 \withappendixtrue\fi
\usepackage{cite}
\usepackage{amsmath,amssymb,amsfonts}
\usepackage{dsfont}
\usepackage{algorithm}
\usepackage{algorithmic}
\usepackage{graphicx}
\usepackage{url}
\usepackage{textcomp}
\usepackage{booktabs}
\usepackage{multirow}
\ifnum\paperversion=1 \usepackage[hidelinks]{hyperref}\fi
\ifnum\paperversion=3 \usepackage[hidelinks]{hyperref}\fi
\def\BibTeX{{\rm B\kern-.05em{\sc i\kern-.025em b}\kern-.08em
    T\kern-.1667em\lower.7ex\hbox{E}\kern-.125emX}}

\newcommand{\spec}{\mathcal{S}}
\newcommand{\voc}{\mathcal{V}}
\newcommand{\fphat}{\hat{\mathbf{f}}}
\newcommand{\fpvec}{\mathbf{f}}
\newcommand{\Enc}{E_{\phi}}

\newcommand{\massvec}{\boldsymbol{\mu}}
\newcommand{\true}{y^{\star}}
\newcommand{\reals}{\mathbb{R}}
\newcommand{\maskt}{\mathsf{m}}

\begin{document}

\title{MARLIN: De Novo Molecular Structure Elucidation from Tandem Mass Spectra without a Ground-Truth Formula}

\ifnum\paperversion=0
\author{\IEEEauthorblockN{Anonymous Submission}
\IEEEauthorblockA{}}
\else
\author{%
\IEEEauthorblockN{Xujun Che$^{1,3}$\quad Xiuxia Du$^{2,3}$\quad Depeng Xu$^{1,3,*}$}
\IEEEauthorblockA{%
$^{1}$Department of Cybersecurity\\
$^{2}$Department of Bioinformatics and Genomics\\
$^{3}$Center for Environmental Monitoring and Informatics Technologies for Public Health\\
University of North Carolina at Charlotte, Charlotte, NC 28223}%
\thanks{This work was made possible by UNC Charlotte internal funding.}%
\thanks{$^{*}$Corresponding author: Depeng Xu (\texttt{dxu7@charlotte.edu}).}%
\versionnotice%
}
\fi

\maketitle

\begin{abstract}
Untargeted tandem mass spectrometry (MS/MS) detects thousands of small molecules per biological sample, yet most go
unidentified because they are absent from spectral libraries. These uncharacterized metabolites and natural products are
precisely the compounds that matter for drug discovery, biomarker research, and exposomics. Computational de novo
structure elucidation could close this gap, but almost all state-of-the-art methods assume the ground-truth molecular
formula is known, an oracle that does not exist for genuinely novel compounds, whose formula must itself be predicted with substantial
error. We present MARLIN, a de novo method that elucidates structures directly from a spectrum with no molecular formula
at any stage. A self-supervised encoder predicts a molecular fingerprint from the raw peaks, and a block-diffusion
language model generates candidate structures conditioned only on the fingerprint and the instrument-measured precursor
mass. A mass-shell constraint, whose heavy-atom prune provably never excludes the true molecule, steers generation toward
the measured mass without fixing the atom inventory, and candidates are accepted by exact parts-per-million mass
agreement. A symmetric noise objective absorbs encoder
error, and a candidate-diversity mechanism keeps the candidates from collapsing to a single structure. On
the NPLIB1 benchmark, MARLIN reaches 16.9\% Top-1 exact-match accuracy, higher than the formula-given
baselines, and returns the closest candidate structures by structural distance and fingerprint similarity. It
recovers the correct molecular formula as a byproduct about as often as a dedicated predictor without ever using one. MARLIN brings de novo structure elucidation to
the discovery regime where the molecular formula is unavailable.
\end{abstract}


\section{Introduction}

\begin{figure*}[t]
\centering
\includegraphics[width=\textwidth]{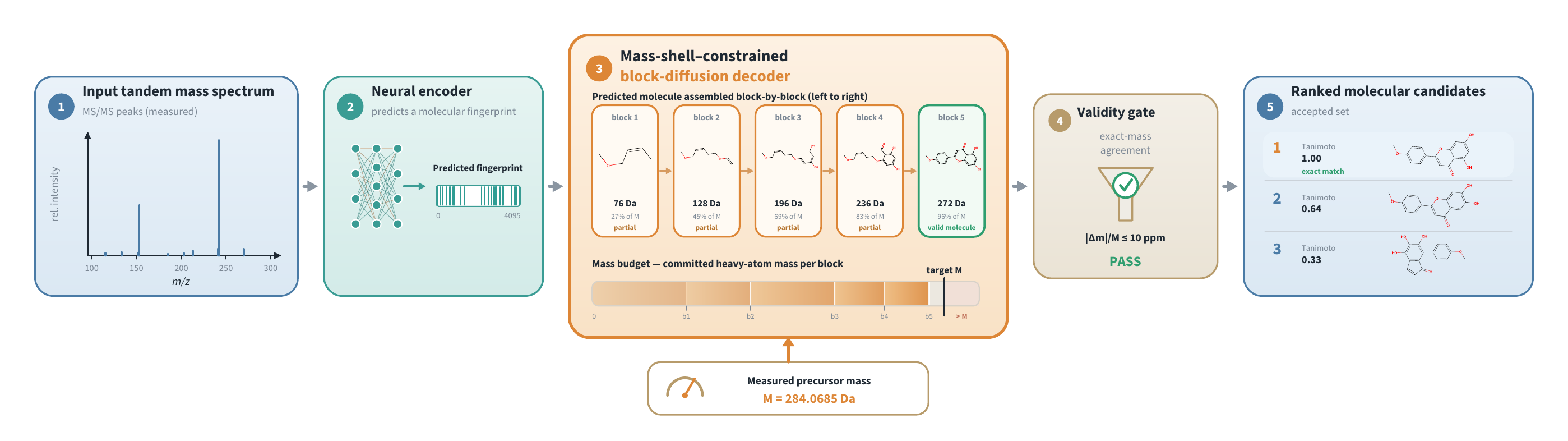}
\caption{Overview of MARLIN, a fully formula-free spectrum-to-structure pipeline, on an NPLIB1 spectrum (acacetin).
(1--2)~A neural encoder turns the spectrum into a predicted Morgan fingerprint. (3)~Conditioned only on that fingerprint
and the measured neutral precursor mass $M$, a mass-shell constrained block-diffusion decoder assembles the molecule
block by block, a provably nondestructive heavy-atom prune (Proposition~1) keeping the committed mass from overshooting
the mass shell. (4)~A validity gate accepts a candidate only when it parses and its exact mass agrees with $M$ within 10 ppm.
(5)~Accepted candidates are ranked by Tanimoto similarity to the predicted fingerprint. The values shown are their
Tanimoto to the reference.}
\label{fig:pipeline}
\end{figure*}

Tandem mass spectrometry (MS/MS) is the workhorse of untargeted metabolomics~\cite{patti2012metabolomics}, natural-product research, and
exposomics~\cite{manz2023nta}, detecting thousands of small molecules per biological sample. Translating a spectrum into a chemical
structure, however, remains the central bottleneck of the field: most detected features go unannotated because they are
absent from spectral libraries~\cite{dasilva2015darkmatter,bittremieux2023suspect,wang2016gnps}, and these
uncharacterized molecules are precisely the novel compounds that matter most for drug
discovery, clinical biomarker studies~\cite{wishart2016metabolomics}, and environmental screening. Computational \emph{structure elucidation}
directly from the spectrum is therefore essential.

Two paradigms dominate. \emph{Database search} predicts a molecular fingerprint from the spectrum and retrieves the
closest known structure, as in CSI:Finger\-ID, SIRIUS, and MIST~\cite{duhrkop2015csifingerid,duhrkop2019sirius,goldman2023mist}.
It is accurate but bounded by database coverage and so cannot, by construction, identify a genuinely novel molecule.
\emph{De novo generation} instead builds a structure from scratch, from early encoder-decoder
models~\cite{litsa2023spec2mol,stravs2022msnovelist,le2020neuraldecipher} to recent deep generative
networks~\cite{bohde2025diffms,sun2026mbgen,qin2026msanchor,wang2025madgen,han2025msbart,bohde2026frigid}, removing the
coverage limit and, in principle, reaching unknown chemistry.

Almost every recent state-of-the-art de novo method, however, shares a hidden assumption: it is given the
\emph{ground-truth molecular formula}~\cite{bohde2025diffms,sun2026mbgen,qin2026msanchor,wang2025madgen,han2025msbart,bohde2026frigid}.
The formula fixes the exact atom inventory (e.g., DiffMS~\cite{bohde2025diffms} generates only the bonds over an atom
set fixed by the formula), acts as a hard candidate filter, supplies a length prior, and re-ranks outputs. This
is an oracle that does not exist in the discovery regime that motivates the task. Formula prediction is itself
error-prone~\cite{che2026comparative,goldman2023mistcf}: on difficult benchmarks the top-1 formula is correct only about
half the time, and a wrong formula propagates directly into a wrong atom set. Evaluating de novo accuracy with the
formula supplied therefore overstates performance on genuinely unknown compounds.

We present \textbf{MARLIN}, a system for the realistic \emph{formula-unknown} setting. Prior methods give the molecular
formula two roles: a conditioning input to the model and the hard constraint on the output atom set. MARLIN hands
both to the measured precursor mass instead. Its \emph{mass-shell constrained block-diffusion decoder} conditions on the
mass together with a predicted fingerprint, and a provably safe heavy-atom prune keeps sampling from overshooting the \emph{mass
shell} set by that mass. Every candidate is accepted only by exact parts-per-million (ppm) mass agreement, never by a
formula. We train the decoder to tolerate realistic encoder error with a symmetric
fingerprint-noise objective and restore the candidate diversity that block decoding otherwise collapses by perturbing the
conditioning per candidate. Fig.~\ref{fig:pipeline} gives an overview of the full pipeline.

MARLIN is \emph{fully formula-free}: a self-supervised spectral encoder
(DreaMS~\cite{bushuiev2026dreams}) predicts the fingerprint directly from the raw peaks, so the entire pipeline, from
spectrum to ranked candidate structures, never computes or consumes a molecular formula. MARLIN is thus directly
applicable in the discovery regime where the formula is genuinely unknown, and it sidesteps the error-prone
formula-prediction step on which prior systems rely. On the NPLIB1 benchmark~\cite{nplib1}, MARLIN achieves strong de
novo accuracy in this harder formula-unknown setting.
Although MARLIN never uses a molecular formula, its exact-mass acceptance makes it recover the correct formula as a
byproduct about as often as a dedicated formula predictor, and more often on the spectra it answers, evidence that a
measured mass and a predicted fingerprint can stand in for an explicit formula. Our contributions are:
\begin{itemize}
\item \textbf{De novo structure elucidation without a ground-truth formula.} We target the realistic formula-unknown
setting, define a clean evaluation contract in which candidates are accepted only by ppm mass agreement, and present
MARLIN, which performs de novo elucidation without any molecular formula.
\item \textbf{A mass-shell constrained block-diffusion decoder.} The decoder generates a variable-length molecule
conditioned on the predicted fingerprint and the measured precursor mass, while a heavy-atom mass prune keeps every
partial molecule from overshooting the mass shell set by that precursor mass. We prove this prune never excludes the true molecule.
\item \textbf{Conditioning diversity for candidate generation.} Decoding each candidate from an independently perturbed
fingerprint produces a diverse candidate set instead of a few repeated structures.
\end{itemize}

\section{Related Work}
\subsection{Fingerprint Prediction and Database Search}
The dominant approach to identifying a molecule from its MS/MS spectrum predicts a molecular fingerprint from the
spectrum and searches a structure database for the closest match. CSI:Finger\-ID established this
spectrum-to-fingerprint-to-database pipeline~\cite{duhrkop2015csifingerid}, SIRIUS couples it with fragmentation-tree
analysis~\cite{duhrkop2019sirius}, and MIST is a strong neural fingerprint predictor~\cite{goldman2023mist}. These
methods are accurate on cataloged compounds but cannot identify a molecule absent from the database, the very case that
matters in discovery. MARLIN reuses a learned fingerprint predictor but \emph{generates} the structure rather than
retrieving it.

\subsection{De Novo Structure Generation from Spectra}
De novo methods synthesize a structure directly from the spectrum. Early systems translate spectra or fingerprints into
molecular strings with encoder-decoder networks (Spec2Mol~\cite{litsa2023spec2mol}, MSNovelist~\cite{stravs2022msnovelist})
or invert fingerprints to structures (Neuraldecipher~\cite{le2020neuraldecipher}). Recent work raises quality with graph
diffusion over a formula-determined atom set (DiffMS~\cite{bohde2025diffms}, MBGen~\cite{sun2026mbgen},
MSAnchor~\cite{qin2026msanchor}, MADGEN~\cite{wang2025madgen}) or with sequence models (MS-BART~\cite{han2025msbart},
FRIGID~\cite{bohde2026frigid}). A common assumption runs through nearly all of this recent work: the ground-truth
molecular formula is given, and it fixes the atom inventory, hard-filters candidates, bounds the output length, and
re-ranks predictions. As formula prediction is itself unreliable~\cite{che2026comparative,goldman2023mistcf}, this assumption
fails in genuine discovery. MARLIN targets the formula-unknown setting: it conditions on the measured precursor mass
rather than a formula and accepts candidates by ppm agreement, so it requires no molecular formula at all.

\subsection{Generative Models for Molecules}
MARLIN's decoder builds on general generative machinery. Molecules are written as SAFE (Sequential Attachment-based Fragment Embedding) strings amenable to left-to-right
generation~\cite{noutahi2024safe}. SAFE reorders SMILES (Simplified Molecular-Input Line-Entry System) strings into chemically meaningful fragments, so committed prefixes can be
parsed as they grow and every token carries a well-defined heavy-atom mass, which is what the mass budget of
Section~\ref{sec:massshell} accumulates. We condition on Morgan fingerprints~\cite{rogers2010ecfp} because they are the
representation that the strongest spectrum-to-fingerprint encoders already predict. Denoising diffusion~\cite{ho2020ddpm} has discrete counterparts for sequences and
graphs (D3PM~\cite{austin2021d3pm}, masked diffusion language models~\cite{sahoo2024mdlm}, and DiGress for
graphs~\cite{vignac2023digress}), and block-diffusion language models interpolate between autoregressive and diffusion
decoding~\cite{arriola2025bd3lm} on a Transformer backbone~\cite{vaswani2017attention}. We adopt a block-diffusion
decoder for its block-by-block, commit-and-freeze sampling. Our contribution is the mass-shell constrained sampler and the
conditioning-diversity mechanism built upon it, not the underlying generative model.

\section{Method}

MARLIN turns a tandem mass spectrum into a ranked list of candidate structures with an \emph{encoder} that predicts a
molecular fingerprint and a \emph{decoder} that generates molecules matching that fingerprint and the measured precursor
mass. Its distinguishing choice is the stoichiometric signal the decoder relies on: prior de novo methods rely on the
\emph{molecular formula} (given or predicted, used both to fix the atom multiset and as a model input) and so inherit
the errors of a formula-prediction step, whereas MARLIN relies only on the \emph{precursor mass} $M$ read from the
instrument, as both the sole stoichiometric input and the sole sampling constraint, with no molecular formula ever used.
We next set up the problem and describe the encoder, the decoder and its conditioning, the sampling-time mass constraint
and its safety guarantee, and two components that make the system effective: noise-robust training and a
candidate-diversity mechanism.

\subsection{Problem Setting and Notation}
We observe a tandem mass spectrum $\spec=\{(m_i,I_i)\}_{i=1}^{P}$ of $P$ peaks with positions $m_i$ and intensities
$I_i$, together with the neutral precursor mass $M = z\,m_{\mathrm{prec}} - \Delta_{a}$, where $z$ is the charge,
$m_{\mathrm{prec}}$ the measured precursor mass-to-charge ratio, and $\Delta_{a}$ the known adduct ion mass. We do \emph{not} observe the molecular formula. The goal is to return a ranked set of candidate molecules
$\{\hat y_1,\dots,\hat y_n\}$ that contains the true molecule $\true$ and ranks it highly.

MARLIN factorizes into an encoder and a decoder. The encoder $\Enc$ maps the spectrum to per-bit fingerprint
probabilities $\fphat=\Enc(\spec)\in[0,1]^{B}$ with $B{=}4096$, which we binarize at threshold $\eta$ to a Morgan
fingerprint~\cite{rogers2010ecfp} $\fpvec=\mathds{1}[\fphat>\eta]\in\{0,1\}^{B}$. We instantiate $\Enc$ with the DreaMS
encoder~\cite{bushuiev2026dreams}, which reads raw peaks only, so no molecular formula enters the pipeline. We
fine-tune the pretrained DreaMS backbone together with a fingerprint head to predict Morgan fingerprints on the NPLIB1
training spectra, augmented with simulated spectra of molecules disjoint from the validation and test sets. The
decoder is a generative model over SAFE~\cite{noutahi2024safe}
token sequences $\mathbf{x}=(x_1,\dots,x_L)\in\voc^{L}$, where $\voc$ is the finite SAFE token vocabulary and $L$ the
sequence length. A deterministic decoding map $g:\voc^{*}\!\to\!\mathcal{M}\cup\{\bot\}$ (with $\voc^{*}$ the set of
token strings of any length) sends a string either to a molecule in the space $\mathcal{M}$ of valid structures or to the
invalid symbol $\bot$ when the string does not decode. Throughout, the decoder is conditioned on $(\fpvec,M)$ and never
on a formula.

\subsection{Conditioning on Mass, Isotopes, and Fingerprint}
The decoder reads a conditioning sequence assembled from measurable quantities,
$Z = [\,\mathbf{z}_{M};\,\mathbf{z}_{\mathrm{iso}};\,\mathbf{z}_{\fpvec}\,]\in\reals^{(2+n_{\mathrm{on}})\times d}$,
injected at every layer by cross-attention, where $d$ is the decoder hidden width. Here
$\mathbf{z}_{M},\mathbf{z}_{\mathrm{iso}}\in\reals^{d}$ are single conditioning tokens for the precursor mass and the
isotope pattern, and $\mathbf{z}_{\fpvec}\in\reals^{n_{\mathrm{on}}\times d}$ is a set of $n_{\mathrm{on}} \leq B$ tokens, one
per on-bit of $\fpvec$. The three are concatenated along the token axis.
The scalar precursor mass is lifted to a vector by a Fourier feature map, $\mathbf{z}_{M}=\mathrm{MLP}(\gamma(M))$ with
$\gamma(M)=[\sin(\omega_r M),\cos(\omega_r M)]_{r=1}^{n_\omega}$ over $n_\omega$ fixed geometrically spaced frequencies
$\omega_r$, so that a single scalar is not ignored by attention. An
optional isotope token $\mathbf{z}_{\mathrm{iso}}$ encodes the $M{+}1$ and $M{+}2$ abundance ratios. It is enabled during
training, where the ratios follow from the structure, and disabled at inference, where they would have to be read from an
MS1 isotope envelope that NPLIB1 does not provide. The fingerprint term $\mathbf{z}_{\fpvec}$ is a set encoding of the
on-bits. The decoder is thus driven by a continuous mass target rather than by an element multiset: where prior de novo decoders
inject a given or predicted molecular formula as a conditioning input, MARLIN injects only the precursor mass, so no
molecular formula enters the conditioning.

\subsection{Block-Diffusion Decoder}
The decoder is a masked discrete-diffusion language model~\cite{ho2020ddpm,austin2021d3pm,sahoo2024mdlm} over SAFE
strings, built on a Transformer~\cite{vaswani2017attention} and made block-causal~\cite{arriola2025bd3lm}. In masked
diffusion, a \emph{forward} process progressively replaces tokens with a special absorbing state $\maskt$ and the model
learns to \emph{reverse} it by predicting the absorbed tokens. Generation then starts from an all-$\maskt$ input and
recovers a sequence by iterative unmasking. Block diffusion applies this reverse process within blocks. We partition the
$L$ positions into $K$ contiguous blocks $\mathcal{B}_1,\dots,\mathcal{B}_K$ of width $w$ and make attention
\emph{block-causal}: positions inside a block attend to one another bidirectionally, but across block boundaries a block
sees only the \emph{earlier} blocks. Within a block the model scores all positions jointly, as in diffusion, while the molecule as a
whole is produced one block at a time from left to right, as in an autoregressive model. This hybrid lets the length
grow during decoding and lets each finished block act as clean left-context for the next.

Concretely, within block $k$ a continuous time $t_k\sim\mathcal{U}(0,1]$ masks each of its tokens independently with
probability $t_k$, and block $k$ is denoised conditioned on the clean prefix $\mathbf{x}_{<k}$ and the conditioning $Z$.
Writing $\mathbf{x}^{t}$ for the noised sequence, the continuous-time absorbing negative evidence lower bound (NELBO) is
\begin{equation}
\mathcal{L}_{\mathrm{dec}}=
\mathbb{E}_{\mathbf{x},Z,k,t_k}
\!\Bigg[\frac{1}{t_k}\!\!\sum_{\substack{j\in\mathcal{B}_k\\ x^{t}_j=\maskt}}\!\!
-\log p_\theta\!\big(x_j\,\big|\,\mathbf{x}^{t}_{\le k},Z\big)\Bigg].
\end{equation}
Here $\mathbf{x}^{t}_{\le k}=(\mathbf{x}_{<k},\mathbf{x}^{t}_{k})$ is the clean committed prefix (blocks $1,\dots,k{-}1$)
concatenated with the noised current block $\mathbf{x}^{t}_{k}$. The inner sum runs over the positions $j\in\mathcal{B}_k$
that the forward process has absorbed to $\maskt$, and the outer expectation draws a block index
$k\sim\mathcal{U}\{1,\dots,K\}$ and its time $t_k\sim\mathcal{U}(0,1]$ jointly with $(\mathbf{x},Z)$ from the training set.
Two mechanisms make this practical. In training, the loss over all $K$ blocks is obtained in a single forward pass over
a \emph{two-stream} input $[\mathbf{x}^{0};\mathbf{x}^{t}]$ under a block-causal mask: the clean stream $\mathbf{x}^{0}$
supplies each block's left-context while the noised stream $\mathbf{x}^{t}$ holds the tokens to be denoised, so the
per-block objectives are computed jointly rather than in a loop. In sampling, the molecule is built one block at a time
from left to right. Each block is initialized fully masked and completed in $w$ denoising steps, each of which re-scores the whole block and
fills one masked position by sampling. A completed block
is then committed and frozen, and its keys and values are cached so later blocks reuse them instead of recomputing.

We adopt this block-by-block scheme because it suits de novo generation on two counts. It is naturally
\emph{variable-length} (blocks are appended until the molecule ends, so the token length need not be fixed in
advance), and because blocks are committed in order, at every step a finalized prefix is available over which the
decoder can track the partial molecule's accumulated mass. That running prefix is precisely what the constraint of the
next subsection uses to hold every candidate to the measured precursor mass, while freezing and caching committed blocks
keeps inference fast.

\subsection{Mass-Shell Constrained Decoding}
\label{sec:massshell}
Nothing in the block-diffusion process ties the growing molecule to the measured precursor mass. Mass-shell constrained
decoding adds this tie at sampling time. Algorithm~\ref{alg:massshell} states the full procedure for generating one
candidate, and the mechanisms below annotate its individual steps. Its goal is to steer every trajectory into the
\emph{mass shell}, the set
of molecules whose total monoisotopic mass lies within a parts-per-million tolerance of $M$, and to do so without ever
committing to a specific element multiset, that is, without a molecular formula. The constraint is applied by masking the
decoder logits at each unmasking step, so it composes directly with the block-wise sampler.

\noindent\textbf{Mass budget.} We precompute a mass table $\massvec\in\reals_{\ge0}^{|\voc|}$ that assigns every token
the summed monoisotopic mass of its \emph{heavy} atoms, excluding hydrogens and setting the special tokens (BOS, EOS,
PAD, MASK) to $0$. We write $\mu_v$ for the entry of $\massvec$ at token $v$. As tokens are committed we accumulate the
committed heavy-atom mass $h=\sum_j\mu_{x_j}$, and, letting $\tau$ denote the mass tolerance in parts per million (ppm),
we write $\delta=\tau\cdot10^{-6}M$ for the tolerance band around the target mass $M$.

\noindent\textbf{Heavy-atom prune.} At each step we set to $-\infty$ the logit of any token whose commitment would push
the heavy-atom mass past the upper edge of the shell (Algorithm~\ref{alg:massshell}, line~\ref{ln:prune}),
\begin{equation}
\text{logit}(v)\leftarrow-\infty \quad\text{if}\quad h+\mu_v>M+\delta. \label{eq:prune}
\end{equation}
Heavy-atom mass is a lower bound on total monoisotopic mass, because the omitted hydrogens can only add mass, so a
partial sequence whose heavy mass already exceeds $M+\delta$ can never be completed on-mass. Pruning these tokens therefore removes
only dead ends. A single scalar comparison enforces the mass constraint without enumerating formulas, and Proposition~1
shows it never removes the true molecule.

\noindent\textbf{Hydrogen-aware valence interval.} Because $\massvec$ omits hydrogens, $h$ \emph{underestimates} the
eventual total mass by the mass of the hydrogens the finished molecule will carry. To reason about the reachable total
mass we bound the hydrogen count from the valence budget of the committed atoms: a connected molecule with $a$ committed
heavy atoms spends at least $2(a-1)$ valence on its heavy-atom skeleton, and the remaining valence is available for
hydrogens. This yields a reachable-mass interval $[\,h,\; h+H_{\max}\,m_{\mathrm H}\,]$, where $H_{\max}$ is the
valence-implied hydrogen capacity (with a slack of four hydrogens, which widens the interval and so relaxes the EOS rule
without excluding any molecule from it) and $m_{\mathrm H}$ is the mass of hydrogen.
The interval sharpens the constraint by telling us, in particular, when a prefix is still too light to reach the
shell even after maximal hydrogen filling.

\noindent\textbf{EOS coupling.} Termination is coupled to this interval so a trajectory neither stops short of the shell
nor overshoots it (Algorithm~\ref{alg:massshell}, line~\ref{ln:eos}). We forbid [EOS] while the molecule is too light, allow it once the
mass can reach the shell, and boost it once even the lightest token would overshoot,
\begin{align}
&\text{forbid [EOS]} \quad\text{if}\quad h+H_{\max}\,m_{\mathrm H}<M-\delta, \nonumber\\
&\text{boost [EOS]} \quad\text{if}\quad h+\mu_{\min}>M+\delta, \nonumber
\end{align}
where $\mu_{\min}$ is the smallest nonzero token mass. Generation thus grows toward $M$ and is allowed to stop as soon
as, but not before, an on-mass structure becomes reachable. The measured mass therefore also takes over the role of the
length prior that a formula supplies to prior decoders.

\noindent\textbf{Grammar and validity masks.} Two further masks keep the support syntactically valid at every step. A
forbidden-token mask removes the structural special tokens, so a position never unmasks to padding, and an incremental
grammar mask removes tokens that would drive the parenthesis balance negative and blocks [EOS] while brackets are open.
Each committed prefix therefore decodes to a well-formed fragment.

\noindent\textbf{Validity gate and acceptance.} Rather than terminating at the first sampled [EOS], we parse as we go:
after each committed block we decode the current prefix with RDKit and finalize the candidate (Algorithm~\ref{alg:massshell},
line~\ref{ln:accept}) the moment it is a valid molecule whose exact, hydrogen-resolved monoisotopic mass satisfies
$|\Delta m|/M\le\tau\cdot10^{-6}$, where $\Delta m = \mathrm{mass}(y)-M$. A trajectory therefore keeps growing until it reaches a structure that is both
valid and mass-exact instead of halting on an invalid or off-mass one, and this ppm test is the exact correctness
criterion applied to every returned candidate. The central guarantee of the heavy-atom prune is the following.

\smallskip
\noindent\textbf{Proposition~1 (Mass-shell safety).}\ \emph{Assume the precursor mass is accurate to the tolerance band,
$|\mathrm{mass}(\true)-M|\le\delta$ with $\delta=\tau\cdot10^{-6}M$. Then, while the SAFE encoding of the true molecule
$\true$ is being decoded, the heavy-atom prune in~\eqref{eq:prune} never masks a correct token.
Constrained decoding therefore cannot eliminate $\true$ through this prune.}

\smallskip
\noindent\emph{Proof.} The token masses $\massvec$ are nonnegative, and the hydrogen atoms omitted from $\massvec$ also
contribute nonnegative mass, so the summed heavy-atom mass of any molecule is a lower bound on its total monoisotopic
mass. Let $x_1,\dots,x_T$ be the tokens of the true SAFE sequence, let $S$ be the set of positions filled so far, each
with its true token, and let $h_S=\sum_{j\in S}\mu_{x_j}$ be the committed heavy-atom mass. Every such partial sequence satisfies
\begin{equation*}
h_S \le \sum_{j=1}^{T}\mu_{x_j} \le \mathrm{mass}(\true) \le M+\delta,
\end{equation*}
where the first inequality adds the unfilled tokens, the second adds the omitted hydrogens, and the third is the tolerance
assumption. When the decoder fills a further position $i\notin S$ with its correct token $x_i$, the same chain applied to
$S\cup\{i\}$ gives $h_S+\mu_{x_i}\le M+\delta$. The prune masks a token $v$ only when $h_S+\mu_v>M+\delta$, so it cannot
mask $x_i$. As this holds at every step, the true molecule remains reachable throughout decoding.~$\square$

\begin{figure}[t]
\centering
\includegraphics[width=0.75\columnwidth]{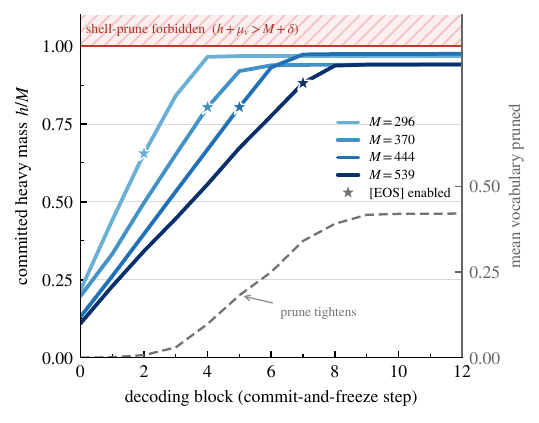}
\caption{Mass-shell constrained decoding on four NPLIB1 spectra. As blocks are committed, the heavy-atom mass $h$ climbs
and plateaus just below $M$. The remaining gap is the hydrogen contribution, resolved at the acceptance step. The prune in~\eqref{eq:prune} forbids tokens that would push $h$ past $M+\delta$ (shaded), and the pruned fraction of the vocabulary
(dashed, right axis) rises as the budget tightens. EOS coupling enables termination (stars) once $h$ can reach $M$.}
\label{fig:masstrace}
\end{figure}

\smallskip
A single measured scalar $M$ therefore steers generation toward the mass shell without ever specifying an element
multiset, and the heavy-atom prune provably preserves the true molecule. The guarantee covers this prune: the
hydrogen-aware interval behind the EOS rule rests on a standard-valence bound with a small slack, and every accepted
candidate is verified on-mass by the final ppm check. Fig.~\ref{fig:masstrace} shows this behavior on real spectra: the
committed heavy-atom mass climbs and plateaus just below $M$, the fraction of pruned tokens rises as the budget
tightens, and EOS coupling enables termination once the mass can reach $M$.

\begin{algorithm}[t]
\caption{Mass-shell constrained block decoding (one candidate)}
\label{alg:massshell}
\textbf{Input}: fingerprint $\fpvec$, precursor mass $M$, tolerance $\tau$, mass table $\massvec$, block width $w$\\
\textbf{Output}: molecule $y$ or $\bot$
\begin{algorithmic}[1]
\STATE $\delta\leftarrow\tau\cdot10^{-6}M$;\ $h\leftarrow 0$;\ prefix $\leftarrow$ [BOS]
\FOR{block $k=1,2,\dots$}
\STATE append $w$ masked tokens
\WHILE{block $k$ has masked positions}
\STATE logits $\leftarrow p_\theta(\cdot\mid \text{prefix}, Z)$
\STATE logit$(v)\!\leftarrow\!-\infty\ \forall v: h+\mu_v>M+\delta$ \label{ln:prune}
\STATE apply EOS coupling and grammar/validity masks \label{ln:eos}
\STATE fill one masked position (sampled), update $h$
\ENDWHILE
\STATE commit/freeze block $k$ (cache clean KV)
\STATE $y \leftarrow g(\text{prefix})$
\IF{$y \neq \bot$ and $|\Delta m|/M \le \tau\cdot10^{-6}$} \label{ln:accept}
\STATE \textbf{return} $y$ \quad(validity gate)
\ENDIF
\IF{[EOS] committed} \STATE \textbf{break} \ENDIF
\ENDFOR
\STATE \textbf{return} $\bot$ \quad(no valid on-mass molecule reached)
\end{algorithmic}
\end{algorithm}

\subsection{Symmetric Fingerprint-Noise Training}
A predicted fingerprint is never exact, so during training we corrupt the conditioning fingerprint \emph{symmetrically}
with probability $p_c$. We drop a fraction $\rho_c\sim\mathcal{U}(\rho_{\min},\rho_{\max})$ of the on-bits and add an
equal number of random off-bits, which gives an expected Tanimoto of $(1-\rho_c)/(1+\rho_c)$. Corrupting in both
directions teaches the decoder to tolerate missing true bits and spurious false bits alike, which matches the error
profile of a real encoder better than dropping bits alone.

\subsection{Conditioning Diversity}
Commit-and-freeze decoding together with the mass-shell constraint makes repeated samples from the same $(\fpvec,M)$
collapse onto one or two reconstructions. To obtain a diverse candidate set, we perturb the \emph{conditioning} rather than
the sampler: candidate $c$ is decoded from $\fpvec^{(c)}=\fpvec\odot\mathbf{b}^{(c)}$, where $\mathbf{b}^{(c)}$ applies an
independent random dropout to the on-bits. Different perturbations commit to different skeletons, so the candidates
cover several distinct structures instead of one or two. We keep the candidates that pass the ppm acceptance test and rank them by Tanimoto
similarity to the predicted fingerprint $\fpvec$.

\section{Experiments}

\subsection{Setup}
We evaluate on the NPLIB1 dataset~\cite{nplib1} for de novo molecular structure generation from MS/MS. Following the standard
protocol~\cite{bohde2025diffms}, we report exact Top-1 and Top-10 \emph{accuracy}, the myopic maximum common edge subgraph distance
(MCES, lower is better), and the Morgan~\cite{rogers2010ecfp} \emph{Tanimoto} similarity
of the Top-1 and Top-10 candidates. The methods fall into two settings. \emph{Formula given}: prior systems that
receive the ground-truth molecular formula, which fixes the atom inventory and/or hard-filters candidates.
\emph{Formula unknown}: FRIGID and MARLIN, which never receive a ground-truth formula. MARLIN uses the fully
formula-free DreaMS~\cite{bushuiev2026dreams} encoder, which takes raw peaks only. Its decoder conditions only on
$(\fpvec,M)$, and candidates are accepted by ppm mass agreement ($\tau{=}10$, a tolerance that covers the mass accuracy
of the Q-TOF and Orbitrap instruments that dominate NPLIB1). FRIGID uses the MIST
encoder~\cite{goldman2023mist}, featurized with a \mbox{MIST-CF}~\cite{goldman2023mistcf} \emph{predicted} top-1 formula, and
its decoder is additionally conditioned on that formula. We evaluate FRIGID-base, without its ICEBERG-guided
inference-time refinement, at the same budget of $384$ decodes per spectrum as MARLIN. It receives the top-1 formula and
spends that budget on it, where the original FRIGID paper's formula-unknown protocol splits the same budget over the
top-5 \mbox{MIST-CF} hypotheses. We retrain \mbox{MIST-CF} on our training split, so no test molecule
is seen in training. It predicts the correct formula for $79.0\%$ of test spectra. All MARLIN and FRIGID numbers are
produced by a single inference pipeline with saved per-spectrum predictions. Table~\ref{tab:hparams} lists the hyperparameters.

\noindent\textbf{Metrics.} Accuracy is computed over all $803$ test spectra, so a spectrum for which a method returns no
candidate counts as a failure. MCES and Tanimoto are averaged over the spectra for which a configuration returns a
candidate. FRIGID always returns one and is evaluated on the spectra MARLIN answers. Their Top-10 values take the best
over the ranked Top-10. Strings that do not parse are never
accepted, and accepted candidates are deduplicated by the connectivity block of the InChIKey, so stereoisomers collapse
to one candidate.

\noindent\textbf{Training data and protocol.} The decoder is trained self-supervised on SAFE strings from the SAFE-GPT
corpus~\cite{noutahi2024safe}: its conditioning (mass, isotope ratios, fingerprint) is computed from each structure with
RDKit, so no spectra are needed, and the conditioning fingerprint is the ground-truth Morgan fingerprint under the
symmetric corruption above. The decoder therefore never sees a spectrum, and any predictor of Morgan fingerprints can
drive it. Molecules whose connectivity InChIKey occurs in the NPLIB1 validation or test splits are excluded from that corpus. The decoder is warm-started from the released FRIGID
decoder, which supplies the Transformer backbone and the fingerprint set-encoder, while the mass and isotope conditioning
modules are newly initialized. Encoders are trained separately and stay frozen during decoder training and at inference.
DreaMS is pretrained self-supervised on unlabeled GNPS~\cite{wang2016gnps} spectra, which may include NPLIB1 test spectra but carry no structures. Its backbone and an iterative (growing) fingerprint head are then fine-tuned together with a cosine loss to the
Morgan fingerprint, at learning rates $10^{-3}$ for the head and $3{\times}10^{-5}$ for the backbone, with
early stopping on validation.
No test molecule shares a connectivity InChIKey with the NPLIB1 training or validation split.

\begin{table}[t]
\caption{MARLIN hyperparameters}
\label{tab:hparams}
\centering
\footnotesize
\setlength{\tabcolsep}{4pt}
\begin{tabular}{ll}
\toprule
Component & Setting \\
\midrule
\multicolumn{2}{l}{\textit{Block-diffusion decoder}} \\
Hidden size / layers / heads & $896$ / $12$ / $14$ \\
SAFE vocabulary & $1880$ \\
Block width $w$ / max length & $8$ / $256$ tokens \\
Denoising steps per block & $8$ (one unmasking per step) \\
\midrule
\multicolumn{2}{l}{\textit{Conditioning}} \\
Morgan fingerprint & $4096$ bits, radius $2$ \\
Binarization threshold $\eta$ (DreaMS) & $0.30$ (tuned on val.) \\
\midrule
\multicolumn{2}{l}{\textit{Decoder training}} \\
Initialization & warm-start from FRIGID \\
Fingerprint noise $p_c$ / $\rho_c$ & $0.5$ / $\mathcal{U}[0.1,0.3]$ \\
Learning rate (AdamW) & $5{\times}10^{-5}$ \\
Batch size / EMA decay & $256$ / $0.9999$ \\
\midrule
\multicolumn{2}{l}{\textit{Mass-shell decoding (inference)}} \\
ppm tolerance $\tau$ & $10$ \\
Hydrogen slack (valence interval) & $4$ hydrogens \\
Decodes per spectrum & $384$ \\
Conditioning-diversity dropout & $0.3$ \\
Softmax temperature & $1.0$ \\
\bottomrule
\end{tabular}
\end{table}

\begin{table*}[!t]
\caption{De novo generation on NPLIB1~\cite{nplib1}. Among formula-unknown methods, the better value per column is
\textbf{bold}; $^\dagger$ marks rows reproduced from DiffMS~\cite{bohde2025diffms}, the rest coming from their original
publications}
\label{tab:main}
\centering
\footnotesize
\begin{tabular*}{\textwidth}{@{\extracolsep{\fill}} l ccc ccc}
\toprule
\multirow{2}{*}{Model} & \multicolumn{3}{c}{Top-1} & \multicolumn{3}{c}{Top-10} \\
\cmidrule(lr){2-4}\cmidrule(lr){5-7}
 & Accuracy $\uparrow$ & MCES $\downarrow$ & Tanimoto $\uparrow$ & Accuracy $\uparrow$ & MCES $\downarrow$ & Tanimoto $\uparrow$ \\
\midrule
\multicolumn{7}{c}{\textit{Formula given (ground-truth formula provided to the model)}} \\
\midrule
Spec2Mol$^\dagger$~\cite{litsa2023spec2mol} & 0.00\% & 27.82 & 0.12 & 0.00\% & 23.13 & 0.16 \\
MIST + Neuraldecipher$^\dagger$~\cite{le2020neuraldecipher} & 2.32\% & 12.11 & 0.35 & 6.11\% & 9.91 & 0.43 \\
MIST + MSNovelist$^\dagger$~\cite{stravs2022msnovelist} & 5.40\% & 14.52 & 0.34 & 11.04\% & 10.23 & 0.44 \\
MADGEN~\cite{wang2025madgen} & 2.10\% & 20.56 & 0.22 & 2.39\% & 12.69 & 0.27 \\
DiffMS~\cite{bohde2025diffms} & 8.34\% & 11.95 & 0.35 & 15.44\% & 9.23 & 0.47 \\
MS-BART~\cite{han2025msbart} & 7.45\% & 9.66 & 0.44 & 10.99\% & 8.31 & 0.51 \\
MSAnchor~\cite{qin2026msanchor} & 8.51\% & 11.12 & 0.38 & 16.90\% & 8.95 & 0.49 \\
MBGen~\cite{sun2026mbgen} & 12.20\% & 7.72 & 0.41 & 22.29\% & 6.71 & 0.50 \\
\midrule
\multicolumn{7}{c}{\textit{Formula unknown (no ground-truth formula)}} \\
\midrule
FRIGID~\cite{bohde2026frigid} & 14.94\% & 10.73 & 0.50 & \textbf{24.03\%} & 8.49 & 0.56 \\
MARLIN & \textbf{16.94\%} & \textbf{6.79} & \textbf{0.55} & 23.54\% & \textbf{5.83} & \textbf{0.60} \\
\bottomrule
\end{tabular*}
\end{table*}

\subsection{Main Results}
Table~\ref{tab:main} compares MARLIN against prior de novo methods on NPLIB1. The formula-given methods are listed in their
\emph{most favorable} configuration, in which the ground-truth molecular formula is handed to the model, fixing the atom
inventory and supporting hard filtering and re-ranking. Even with that oracle, the formula-given baselines in
Table~\ref{tab:main} reach at most $12.20\%$ Top-1 accuracy (MBGen), which MARLIN exceeds without ever using a formula,
at $16.94\%$. Re-evaluating these
baselines in the formula-unknown
setting is feasible, by supplying a \mbox{MIST-CF} predicted top-1 formula exactly as we do for FRIGID, but it would only lower
them further, so we report each in its favorable formula-given configuration and evaluate MARLIN and FRIGID without any
ground-truth formula.

Among the formula-unknown methods, MARLIN leads in Top-1 accuracy ($16.94\%$ vs.\ $14.94\%$) and in every structural
metric, while FRIGID is slightly ahead at Top-10 ($24.03\%$ vs.\ $23.54\%$). MARLIN is fully formula-free, with no
formula predicted or supplied anywhere in the pipeline, whereas FRIGID is conditioned on a predicted formula. On the
same spectra, MARLIN's Top-1 MCES is $6.79$ against $10.73$ for FRIGID and its Top-1 Tanimoto $0.55$ against $0.50$, so
even when it misses the exact structure its top candidate stays close to the reference.
\ifwithappendix Appendix~\ref{app:mist} also reports MARLIN with the MIST encoder that FRIGID uses.\fi

\subsection{Ablations}
We ablate MARLIN in Table~\ref{tab:abl}. Removing \emph{conditioning diversity}, the per-candidate fingerprint perturbation, causes the largest single drop
(Top-1 $16.94\%\!\to\!12.83\%$, Top-10 $23.54\%\!\to\!14.94\%$). The cause is intrinsic to block decoding: once a block
is committed and frozen, repeated draws from the same $(\fpvec,M)$ retrace the same skeleton, so the candidate set collapses. Without the perturbation, the median answered spectrum returns two distinct structures among its ranked
Top-10, against six with it, and Top-10 adds little over Top-1. Decoding each candidate from an independently perturbed fingerprint prevents this collapse. The same perturbation does little for FRIGID: applied there it leaves Top-1 unchanged ($14.94\%$), moves
Top-10 only slightly ($24.03\%\!\to\!24.53\%$), and lowers Tanimoto ($0.50\!\to\!0.44$), because FRIGID's full-sequence sampler already yields diverse candidates, so the perturbation mainly
discards useful signal (Table~\ref{tab:abl}, lower group). The collapse is thus
specific to the block-diffusion decoder, the same design that affords the mass-shell budget and key/value caching, and
conditioning diversity is the fix it needs. Removing \emph{mass conditioning} together with the heavy-atom prune, so that the decoder is conditioned on the
fingerprint alone, costs accuracy
($16.94\%\!\to\!14.69\%$): the precursor mass carries usable signal. Removing only the \emph{heavy-atom prune}
while keeping mass conditioning barely changes accuracy ($16.94\%\!\to\!16.44\%$). The prune is a feasibility mechanism,
not an accuracy lever: it steers decoding toward on-mass candidates, while EOS coupling and the final ppm acceptance,
which this ablation keeps, still hold every returned candidate to the measured mass. Noise-augmented
decoder training (symmetric fingerprint corruption) helps at a matched training step ($13.20\%\!\to\!15.19\%$ Top-1).
Some lower-accuracy configurations reach a higher Tanimoto (fingerprint only peaks at $0.62$ at Top-1) because they
answer fewer spectra, over which their Tanimoto is averaged, and keep fewer but structurally sharper candidates.

\begin{table}[!t]
\caption{Ablation studies on NPLIB1}
\label{tab:abl}
\centering
\footnotesize
\setlength{\tabcolsep}{2pt}
\begin{tabular}{l cc cc}
\toprule
\multirow{2}{*}{Configuration} & \multicolumn{2}{c}{Top-1} & \multicolumn{2}{c}{Top-10} \\
\cmidrule(lr){2-3}\cmidrule(lr){4-5}
 & Acc.\ $\uparrow$ & Tanimoto $\uparrow$ & Acc.\ $\uparrow$ & Tanimoto $\uparrow$ \\
\midrule
\multicolumn{5}{l}{\textit{MARLIN: remove components}} \\
\midrule
full & \textbf{16.94\%} & 0.55 & \textbf{23.54\%} & 0.60 \\
\quad w/o conditioning diversity & 12.83\% & 0.57 & 14.94\% & 0.59 \\
\quad w/o mass cond.\ \& prune & 14.69\% & 0.62 & 19.18\% & 0.66 \\
\quad w/o heavy-atom prune & 16.44\% & 0.56 & 23.04\% & 0.60 \\
\midrule
\multicolumn{5}{l}{\textit{Decoder training (matched step)}} \\
\midrule
without noise augmentation & 13.20\% & 0.54 & 16.19\% & 0.58 \\
with noise augmentation & 15.19\% & 0.57 & 20.42\% & 0.61 \\
\midrule
\multicolumn{5}{l}{\textit{Conditioning diversity applied to FRIGID}} \\
\midrule
FRIGID & 14.94\% & 0.50 & 24.03\% & 0.56 \\
\quad + conditioning diversity & 14.94\% & 0.44 & 24.53\% & 0.53 \\
\bottomrule
\end{tabular}
\end{table}

\begin{figure}[t]
\centering
\includegraphics[width=\linewidth]{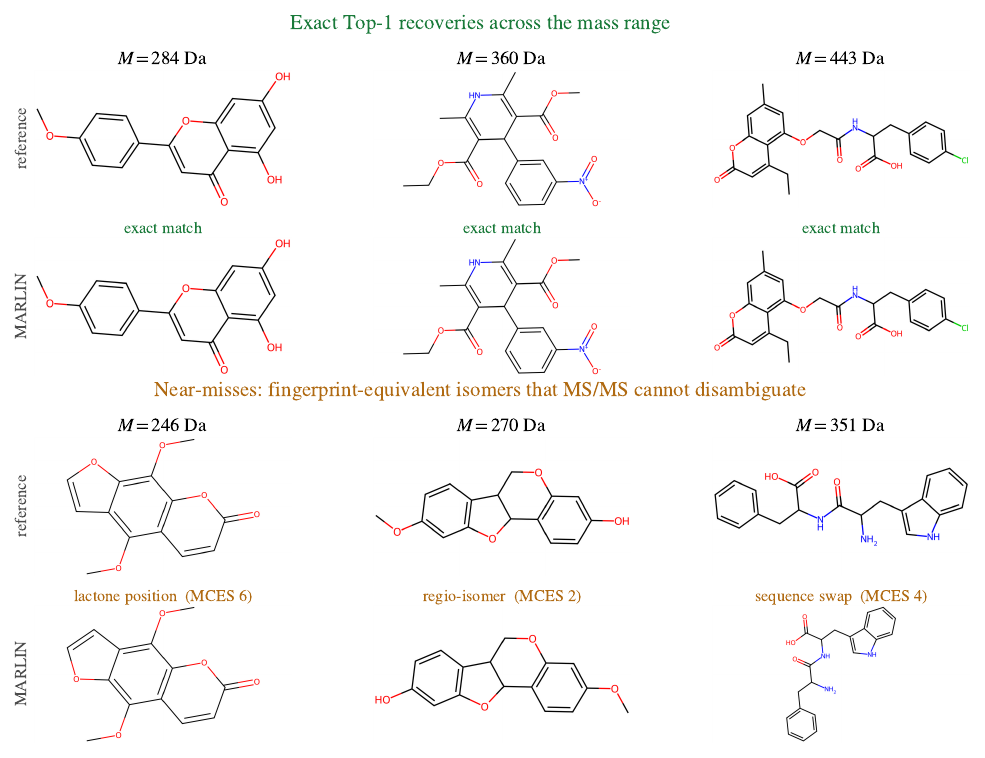}
\caption{Qualitative MARLIN predictions with the DreaMS encoder. Top: exact Top-1 recoveries, from a flavone ($284$ Da) to a chlorinated coumarin conjugate ($443$ Da), with the
MARLIN structure identical to the reference. Bottom: representative near-misses, where the Top-1 candidate is a distinct
isomer that is nearly fingerprint-equivalent to the reference (Tanimoto near $1$, MCES $2$ to $6$). The differences
are a lactone position, a ring regiochemistry, and a dipeptide sequence permutation, all classic
ambiguities that the predicted fingerprint does not resolve. Such errors explain why MARLIN attains high structural
similarity even when an exact match is missed.}
\label{fig:examples}
\end{figure}

\subsection{Further Analysis}
\label{sec:analysis}
\textbf{Qualitative examples.} Fig.~\ref{fig:examples} shows representative DreaMS predictions. MARLIN recovers diverse scaffolds exactly, and its characteristic failure mode is informative: in the closest cases the
Top-1 candidate is a distinct isomer that is nearly fingerprint-equivalent to the reference, differing only by a
positional detail such as a lactone or ring regiochemistry, a double-bond position, or a residue order. These are
distinctions the predicted fingerprint does not capture, so they bound exact-match accuracy while the predicted
structure stays close to the reference. This is the fingerprint bottleneck that the error analysis below quantifies, seen at
the level of individual molecules.

\textbf{Error analysis.} The model rarely returns a wholly unrelated structure. Across the nonexact Top-1 predictions,
the candidate remains in the structural neighborhood of the reference, with a median MCES of $6$ and roughly $70\%$
within an MCES of $10$, and $96.0\%$ of these predictions carry the true molecular formula. The errors are therefore
almost entirely isomers of the reference, the closest being the near-identical ones in Fig.~\ref{fig:examples}: the
measured mass settles the composition, and what is left to decide is which isomer explains the spectrum. That decision
rests on the encoder fingerprint, not the decoder: the predicted fingerprints average a Tanimoto of $0.55$ to the true
Morgan fingerprint, so the decoder reconstructs from a substantially corrupted bit vector.

\textbf{Formula recovery.} MARLIN never uses a molecular formula, yet its Top-1 candidate carries the true one for
$76.7\%$ of all spectra and $96.9\%$ of the spectra it answers. The ppm acceptance test is what makes this work: it admits
only candidates whose exact mass matches the precursor, which leaves a few compatible formulas for the predicted
fingerprint to choose among. The mass components instead govern how often a candidate is produced at all, and recovery falls to $54.0\%$ once mass conditioning and the prune are removed. A dedicated predictor, \mbox{MIST-CF} retrained on the same split, reaches
$79.0\%$ over the same $803$ spectra and $83.3\%$ on the spectra MARLIN answers, where MARLIN reaches $96.9\%$. The two
disagree on $108$ of those spectra and MARLIN is right on $97$ of them. Narrowing to the few mass-compatible formulas
and then choosing among them with a predicted fingerprint brings structural evidence to bear that a formula-only
predictor cannot use. FRIGID's Top-1 candidate carries the true formula for $33.6\%$ of all
spectra in the same formula-unknown setting.

\textbf{Performance by molecular mass.} Table~\ref{tab:massbin} breaks Top-1 accuracy down by neutral precursor mass into
three bins ($<$300, 300--500, and $\ge$500 Da, holding 259, 395, and 149 spectra respectively).
The $\ge$500 Da regime is the hardest for both methods, consistent with the larger search
space (and, for MARLIN, the looser hydrogen budget). FRIGID is more accurate on small molecules
($<$300 Da), whereas MARLIN is more accurate above 300 Da ($18.7\%$ vs.\ $14.2\%$ at 300--500 Da and $11.4\%$ vs.\
$7.4\%$ at $\ge$500 Da). The sharp drop above 500 Da motivates a larger-molecule decoder as
future work.

\begin{table}[t]
\caption{Top-1 accuracy by neutral precursor mass (Da)}
\label{tab:massbin}
\centering
\footnotesize
\setlength{\tabcolsep}{6pt}
\begin{tabular}{l ccc}
\toprule
Method & $<$300 & 300--500 & $\ge$500 \\
\midrule
FRIGID & 20.5\% & 14.2\% & 7.4\% \\
MARLIN & 17.4\% & 18.7\% & 11.4\% \\
\bottomrule
\end{tabular}
\end{table}

\textbf{Decoding efficiency.} The block-diffusion decoder is also faster than the full-sequence diffusion decoder it
replaces. Given the same fingerprints and a matched budget of $384$ decodes per spectrum, on a single NVIDIA L40S,
MARLIN's decoder takes $34.1$ s per spectrum on average against $55.8$ s for the full-sequence FRIGID decoder, a
$1.6\times$ speedup that isolates the decoder. Block-causal decoding caches the keys and values of committed blocks instead of recomputing
the whole sequence, and the validity gate often finalizes a candidate before the maximum length, so the per-spectrum cost
is lower despite the added mass-shell bookkeeping. These times measure the decoder alone. FRIGID additionally needs
the \mbox{MIST-CF} formula prediction and the peak-to-subformula assignment that its MIST encoder requires, a substantial
additional cost, whereas MARLIN's DreaMS encoder reads raw peaks and avoids this dependency entirely, so the end-to-end
efficiency advantage is larger than the decoder-only comparison suggests.

\section{Conclusion}
We presented MARLIN, a system for de novo molecular structure elucidation from tandem mass spectra that operates without
a ground-truth molecular formula. MARLIN predicts a molecular fingerprint from the spectrum and generates structures with
a mass-shell constrained block-diffusion decoder conditioned on the fingerprint and the measured precursor mass,
accepting candidates by exact ppm mass agreement. The mass-shell constraint steers generation on-mass without an element
multiset, and its heavy-atom prune provably never excludes the true molecule. A self-supervised spectral encoder lets the entire pipeline
run with no formula anywhere. On the NPLIB1 benchmark MARLIN reaches 16.9\% Top-1 exact-match accuracy, higher than the
formula-given baselines. By elucidating structures
directly from the spectrum, MARLIN addresses a practical bottleneck in untargeted metabolomics and natural-product
discovery, where the molecular formula is often unavailable. Promising directions include scaling the mass-shell decoder
to larger molecules through a hydrogen- and formula-aware mass budget, and further improving encoder fingerprint quality,
which most directly governs end-to-end accuracy.

\section*{Code and Data Availability}
All experiments use the public NPLIB1 benchmark. Code and trained models are available at\newline
\url{https://github.com/du-lab-data-science/MARLIN}.

\bibliographystyle{IEEEtran}
\bibliography{marlin}

\ifwithappendix
\appendices
\section{MARLIN with the MIST Encoder}
\label{app:mist}
Besides the fully formula-free DreaMS encoder, we evaluate MARLIN with the MIST encoder~\cite{goldman2023mist} that
FRIGID uses (its released checkpoint, not fine-tuned by us), $\fphat=\Enc(\spec,\tilde F)$. MIST featurizes each peak with subformulae of a formula $\tilde F$. We
supply the \mbox{MIST-CF}~\cite{goldman2023mistcf} predicted top-1 formula, used only to compute these peak features. The
decoder is unchanged: it conditions only on $(\fpvec,M)$, never sees $\tilde F$, and accepts candidates by ppm mass
agreement. Because its encoder consumes a predicted formula, this configuration is not formula-free, but it shares
FRIGID's encoder and predicted formula, so the comparison with FRIGID isolates the decoder.

Table~\ref{tab:mist} compares both MARLIN configurations with FRIGID. So that structural metrics are not compared over
different sets of spectra, FRIGID's MCES and Tanimoto are computed on exactly the spectra that the respective MARLIN
configuration answers. With the shared encoder, MARLIN (MIST) reaches $19.18\%$
Top-1 and $26.65\%$ Top-10 against $14.94\%$ and $24.03\%$ for FRIGID. On the spectra it answers, its Top-1 MCES is
lower than FRIGID's ($8.59$ vs.\ $8.86$), whereas FRIGID has the higher Tanimoto and the lower Top-10 MCES there. The
clear structural advantage of MARLIN comes with the DreaMS encoder.

\begin{table}[!h]
\caption{Both MARLIN encoders against FRIGID. In each group, MCES and Tanimoto are computed on the spectra that
MARLIN configuration answers; the better value in each pair is \textbf{bold}}
\label{tab:mist}
\centering
\footnotesize
\setlength{\tabcolsep}{2.5pt}
\begin{tabular}{l ccc ccc}
\toprule
\multirow{2}{*}{Model} & \multicolumn{3}{c}{Top-1} & \multicolumn{3}{c}{Top-10} \\
\cmidrule(lr){2-4}\cmidrule(lr){5-7}
 & Acc.\,$\uparrow$ & MCES\,$\downarrow$ & Tani.\,$\uparrow$ & Acc.\,$\uparrow$ & MCES\,$\downarrow$ & Tani.\,$\uparrow$ \\
\midrule
\multicolumn{7}{l}{\textit{Spectra answered by MARLIN (DreaMS)}} \\
FRIGID & 14.94 & 10.73 & 0.50 & \textbf{24.03} & 8.49 & 0.56 \\
MARLIN (DreaMS) & \textbf{16.94} & \textbf{6.79} & \textbf{0.55} & 23.54 & \textbf{5.83} & \textbf{0.60} \\
\midrule
\multicolumn{7}{l}{\textit{Spectra answered by MARLIN (MIST)}} \\
FRIGID & 14.94 & 8.86 & \textbf{0.53} & 24.03 & \textbf{6.77} & \textbf{0.59} \\
MARLIN (MIST) & \textbf{19.18} & \textbf{8.59} & 0.51 & \textbf{26.65} & 7.40 & 0.57 \\
\bottomrule
\end{tabular}
\end{table}

By neutral precursor mass, MARLIN (MIST) reaches $27.8\%$, $17.2\%$, and $9.4\%$ Top-1 accuracy in the $<$300,
300--500, and $\ge$500 Da bins (FRIGID: $20.5\%$, $14.2\%$, and $7.4\%$), so it is strongest on small molecules. Its
Top-1 candidate carries the true molecular formula for $72.2\%$ of all spectra. Ablating the mass components with this
encoder follows the DreaMS pattern: removing mass conditioning and the prune lowers accuracy (Top-1 $19.18\%\!\to\!18.93\%$, Top-10
$26.65\%\!\to\!24.66\%$) while raising the Top-1 Tanimoto to $0.61$, and removing only the heavy-atom prune does not lower accuracy ($20.05\%$ Top-1, $28.14\%$ Top-10), in line with its role as a feasibility mechanism rather than an accuracy lever. The decoding times in Section~\ref{sec:analysis} were measured with this
encoder's fingerprints for both decoders.
\fi

\end{document}